\documentclass[letterpaper, 10 pt, conference]{ieeeconf} 
\IEEEoverridecommandlockouts

\usepackage{cite}
\usepackage{amsmath,amssymb,amsfonts}
\usepackage{algorithmic}
\usepackage{graphicx}
\usepackage{textcomp}
\usepackage{xcolor}
\usepackage{subfigure}
\usepackage{hyperref}

\def\BibTeX{{\rm B\kern-.05em{\sc i\kern-.025em b}\kern-.08em
    T\kern-.1667em\lower.7ex\hbox{E}\kern-.125emX}}
\begin{document}


\title{\LARGE \bf
Universal Semantic Segmentation for Fisheye Urban Driving Images
}

\author{Yaozu Ye$^{1}$, Kailun Yang$^{2}$, Kaite Xiang$^{1}$, Juan Wang$^{1}$ and Kaiwei Wang$^{3}$
\thanks{$^{1}$Y. Yao, K. Xiang and J. Wang are with State Key Laboratory of Modern Optical Instrumentation, Zhejiang University, China {\tt \{yaozuye, katexiang, zjuwjopt\}@zju.edu.cn}}
\thanks{$^{2}$K. Yang is with Institute for Anthropomatics and Robotics, Karlsruhe Institute of Technology, Germany {\tt kailun.yang@kit.edu}}
\thanks{$^{3}$K. Wang is with National Optical Instrumentation Engineering Technology Research Center, Zhejiang University, China {\tt wangkaiwei@zju.edu.cn}}
}

\maketitle

\begin{abstract}
Semantic segmentation is a critical method in the field of autonomous driving. When performing semantic image segmentation, a wider field of view (FoV) helps to obtain more information about the surrounding environment, making automatic driving safer and more reliable, which could be offered by fisheye cameras. However, large public fisheye datasets are not available, and the fisheye images captured by the fisheye camera with large FoV comes with large distortion, so commonly-used semantic segmentation model cannot be directly utilized. In this paper, a seven degrees of freedom (DoF) augmentation method is proposed to transform rectilinear image to fisheye image in a more comprehensive way. In the training process, rectilinear images are transformed into fisheye images in seven DoF, which simulates the fisheye images taken by cameras of different positions, orientations and focal lengths. The result shows that training with the seven-DoF augmentation can improve the model’s accuracy and robustness against different distorted fisheye data. This seven-DoF augmentation provides a universal semantic segmentation solution for fisheye cameras in different autonomous driving applications. Also, we provide specific parameter settings of the augmentation for autonomous driving. At last, we tested our universal semantic segmentation model on real fisheye images and obtained satisfactory results. The code and configurations are released at \url{https://github.com/Yaozhuwa/FisheyeSeg}.
\end{abstract}

\section{Introduction}
With the research boom of autonomous driving, scene understanding has become a hot topic. Semantic segmentation enables pixel-by-pixel tagging of images, completing several fine detection tasks at the same time, which makes it ideal for intelligent vehicles, advanced driver assistance systems, as well as personal wearable navigation tools~\cite{yang2018unifying}\cite{wang2018environmental}.

Thanks to the emergence of large-scale natural datasets~\cite{cordts2016cityscapes}\cite{huang2018apolloscape} and architectural advances of convolutional neural networks~\cite{long2015fully}\cite{orsic2019defense}, most current semantic segmentation studies are based on images taken by pinhole cameras. The urban traffic environment is so complex that more information of the surroundings is required, while the pinhole camera only has a narrow FoV. If a vehicle or pedestrian suddenly appears from the blind spot, the safety of autonomous driving is difficult to guarantee. One of the approach is to increase the amount of information obtained for complete scene comprehension. 

In this line, panoramic camera and multi-sensor fusion are good solutions~\cite{yang2019can}\cite{yang2019pass}. For example, by installing multiple cameras on the vehicle, or attaching additional ultrasonic radar and LiDAR sensors, we can increase the amount of acquired information~\cite{madawy2019rgb}\cite{yang2019ds}. However, these methods entail additional semantic mapping and fusion of data from multiple sensors, where repeated calibration and matching are required for different designed hardware collocation methods~\cite{long2018fusion}. A vital subset of systems~\cite{narioka2018understanding}\cite{wu2018vh}\cite{deng2019restricted} mapped semantic segmentation from a stack of surround-view images into a bird-eye space, which works well for estimating road layout but sacrifices safety-critical view above the horizon. Nevertheless, generating on the fly a holistic representation incurs significant latency and computational cost to process multi-view images. In addition, the multi-sensor approach is cumbersome and very expensive, which could be problematic in some application scenarios~\cite{yang2019can}. A simpler and more direct way is to utilize a fisheye camera which naturally images a wide FoV~\cite{yeol2018scene}.

While the fisheye image has a large FoV, it has large distortion. The distortion of the object depends on the view angle of the object relative to the fisheye camera. Moreover, the distortion of fisheye lenses with different focal lengths are also different. Due to the existence of distortion, the commonly-used semantic segmentation model for pinhole camera cannot be directly applied in fisheye image segmentation. There are two solutions to it. One is to rectify the fisheye image to rectilinear image, followed by common segmentation methods to obtain the wide-FoV semantic map. However, the rectification process will lead to a loss of boundary information. The output image will have a smaller FoV, which is against the original intention of using a fisheye camera.

The second approach is to carry out the segmentation directly on the fisheye images. This approach works on condition that we have a large-scale finely-annotated fisheye dataset to train our model. However, currently there is not a fisheye image dataset that fits exactly the purpose, while collecting and annotating such a dataset is expensive and laborious. WoodScape~\cite{yogamani2019woodscape}, a multi-task, multi-camera fisheye dataset for autonomous driving, and OmniScape~\cite{sekkat202omniscape}, a synthetic dataset, could  meet our requirement, but they have not been released yet. In addition, they are not as diverse and realistic as large-scale pinhole datasets~\cite{cordts2016cityscapes}\cite{huang2018apolloscape}, while the large gap between synthetic and real-world domain necessitates further domain adaptation strategies. Even if we get it, there are also many problems to resolve. For example, how can we apply the model to a different fisheye camera (for they have different distortion). And is the model fits the situation that a fisheye camera installed on a different height and orientation.

Some other researches begin to train their model on synthetic fisheye dataset. The early research~\cite{deng2017cnn} relied on a $r=f\theta$ model to transform the rectilinear images to fisheye images (which they called it zoom augmentation), and trained their model with the synthetic dataset. This zoom augmentation deals the lack of the fisheye dataset to some extent, but the synthetic dataset is not rich enough. The follow-up researches~\cite{saez2018cnn}\cite{saez2019real} inherited the zoom augmentation to synthesis virtual fisheye dataset, but focused on the CNN (Convolutional Neural Networks) structure design. Blott et al.~\cite{blott2018semantic} proposed to use the projection model transformation (PMT) to synthesize fisheye data, which they called it six-DoF (six degrees of freedom) augmentation. This six-DoF augmentation simulated the situation that camera rotated (three DoF) and shifted (three DoF) with respect to the coordinate system origin and axes. It gives a better way to generate fisheye dataset. However the research aimed at general semantic segmentation and didn't realize the great potential of the six-DoF method in the field of autonomous driving. Another work~\cite{qian2019oriented} on pedestrian detection also utilized the PMT to generate their fisheye data, but they only explicated the camera rotation around the vertical line (one DoF). To deal with the distortion of fisheye images, Deng et at.~\cite{deng2019restricted} designed a CNN structure based on deformable convolution~\cite{dai2017deformable} and obtained a better performance. 

A big obstacle to the practical application of semantic segmentation in automatic driving is the robustness of semantic segmentation algorithm, which requires high segmentation accuracy in different scenarios. In particular, the robustness of the algorithm is critical for the safety requirements of automatic driving. The robustness of the algorithm not only depends on the design of CNN network structure, but also largely depends on the dataset we feed it. 

In this paper, we focus on the data data augmentation method to transform the rectilinear dataset to synthetic fisheye dataset. Based on the six-DoF augmentation, we desgined seven fisheye augmentation methods and tested them on testing sets of different distortion, and the results showed that the seven-DoF method have the best generalization capacity. So we proposed the seven-DoF augmentation method and apply it to fisheye urban driving images. This proposed seven-DoF augmentation transforms rectilinear images to virtual fisheye images taken by fisheye cameras with different angles, positions and distortion parameters during training, which is perfectly suitable for the urban driving images. It provides a universal semantic segmentation solution for fisheye cameras in different autonomous driving applications. Also, we conduct extensive experiment to investigate the hyper-parameters settings for the seven-DoF data augmentation and give specific values of our hyper-parameters settings.

In the method section, we explain the principle and intuitive understanding of the seven-DoF augmentation in detail. In the experiments section, we conduct several experiments to prove the superiority of the seven-DoF augmentation. Also, the setting of hyper-parameters for data augmentation is discussed there. Next, we test our universal semantic segmentation model on real fisheye images which are captured by a smart phone with an external fisheye lens and get satisfactory results. At last, we make a summary of the article and suggest some possible future research directions of fisheye segmentation.

\begin{figure}[tb]
    \centering
    \includegraphics[width=0.4\textwidth]{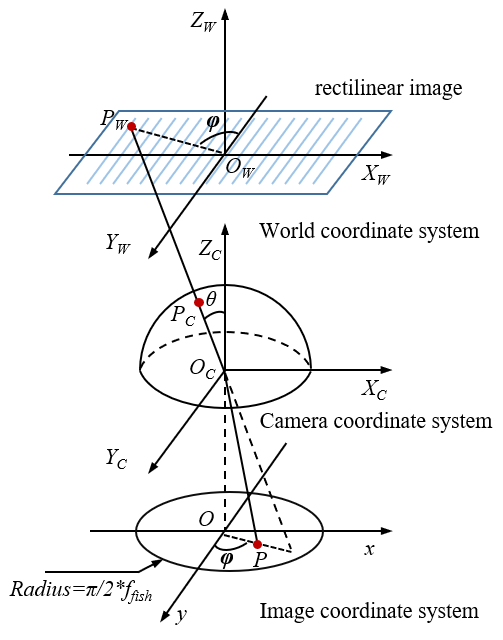}
    \caption{Projection model of fisheye camera. $P_W$ is a point on a rectilinear image that we place on the x-y plane of the world coordinate system. $\theta$ is the Angle of incidence of the point relative to the fisheye camera. $P$ is the imaging point of $P_W$ on the fisheye image. $|OP| = f \theta$. The relative rotation and translation between the world coordinate system and the camera coordinate system results in six degrees of freedom.}
    \label{fig:PMT}
\end{figure}

\begin{figure*}[t]
  \centering
    \subfigure[Camera moves to the left (X)]
    {\includegraphics[width=0.21\textwidth]{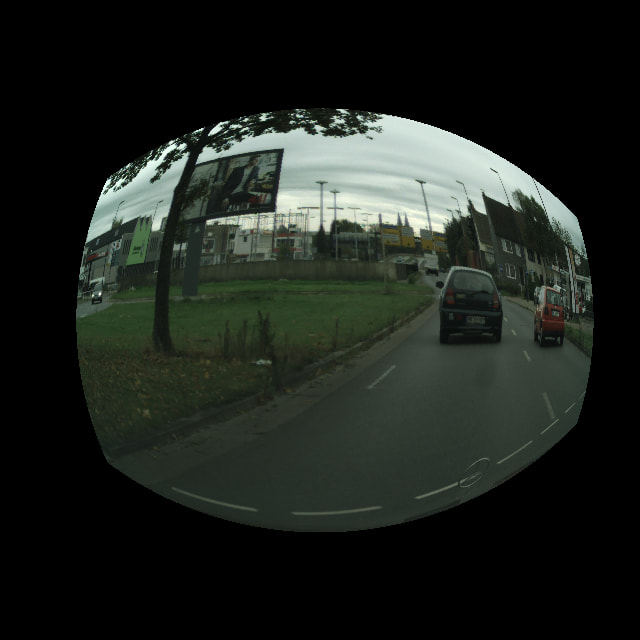}
    \label{fig:trans:left}} 
    \subfigure[Camera moves to the right (X)]
    {\includegraphics[width=0.21\textwidth]{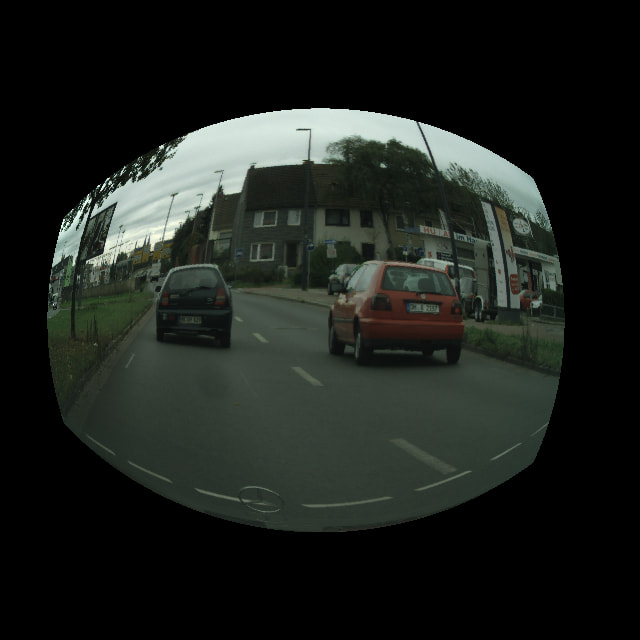}
    \label{fig:trans:right}} 
    \subfigure[Camera moves up (Y)]
    {\includegraphics[width=0.21\textwidth]{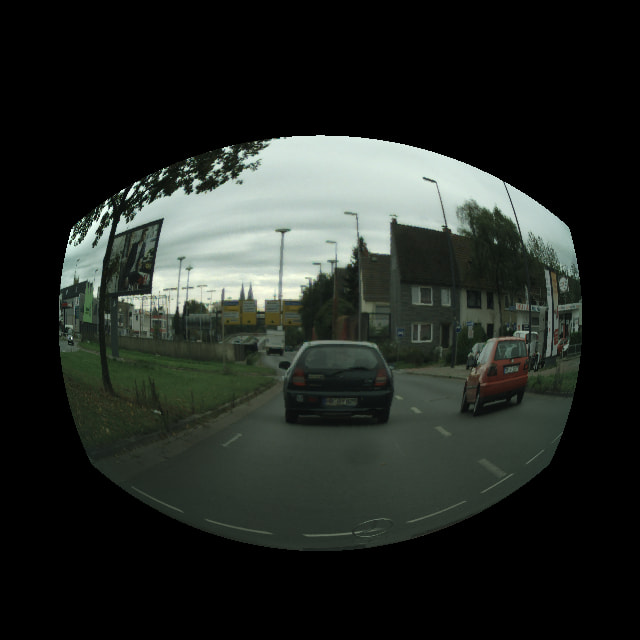}
    \label{fig:trans:up}} 
    \subfigure[Camera moves down (Y)]
    {\includegraphics[width=0.21\textwidth]{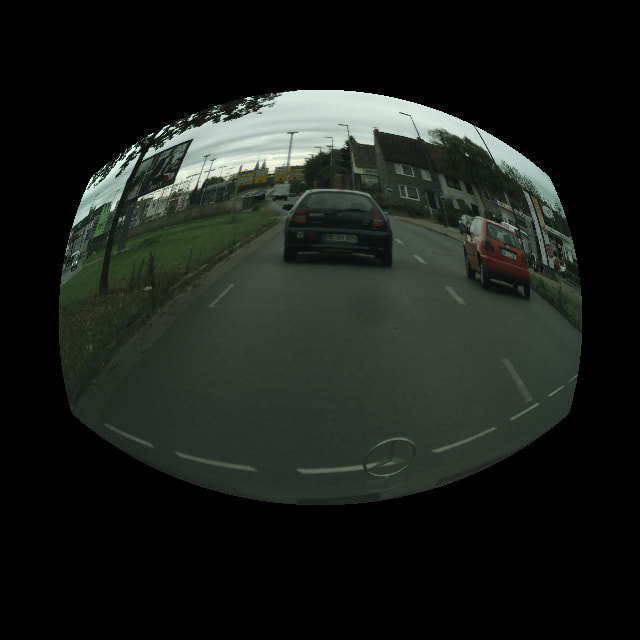}
    \label{fig:trans:down}} \\

    \subfigure[Camera moves forward (Z)]
    {\includegraphics[width=0.21\textwidth]{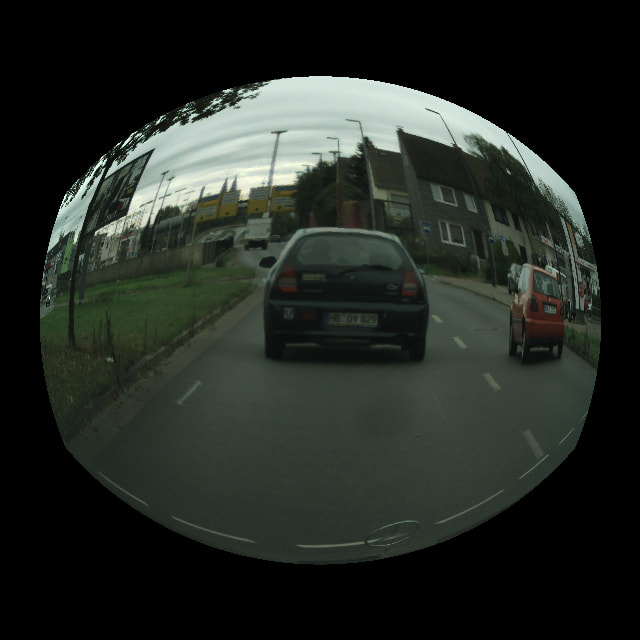}
    \label{fig:trans:forward}}
    \subfigure[Camera moves back (Z)]
    {\includegraphics[width=0.21\textwidth]{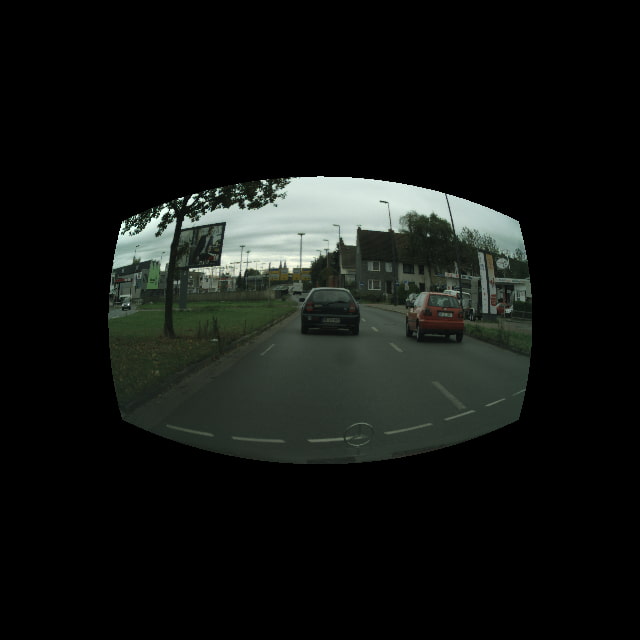}
    \label{fig:trans:back}} 
    \subfigure[Camera turns left (Y)]
    {\includegraphics[width=0.21\textwidth]{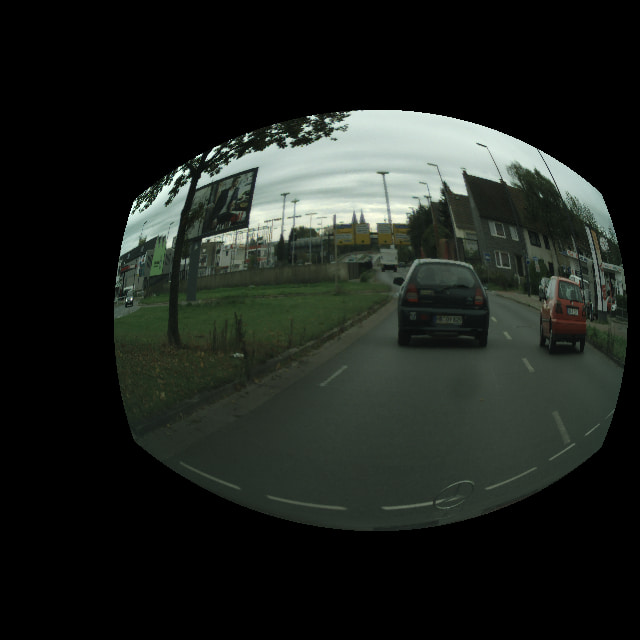}
    \label{fig:rotate:left}} 
    \subfigure[Camera turns right (Y)]
    {\includegraphics[width=0.21\textwidth]{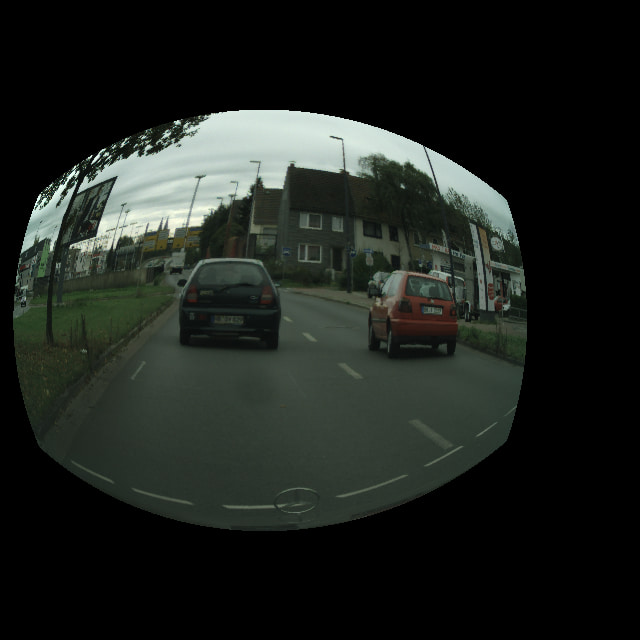}
    \label{fig:rotate:right}} \\
    
    \subfigure[Camera turns up (X)]
    {\includegraphics[width=0.21\textwidth]{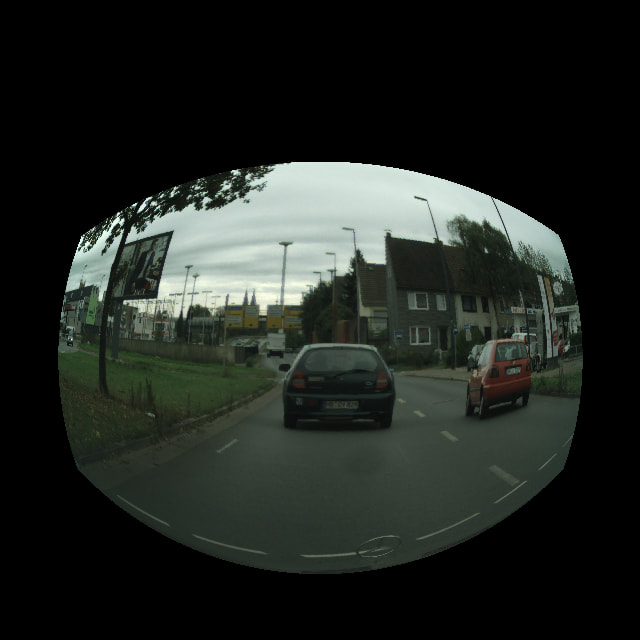}
    \label{fig:roate:up}}
    \subfigure[Camera turns down (X)]
    {\includegraphics[width=0.21\textwidth]{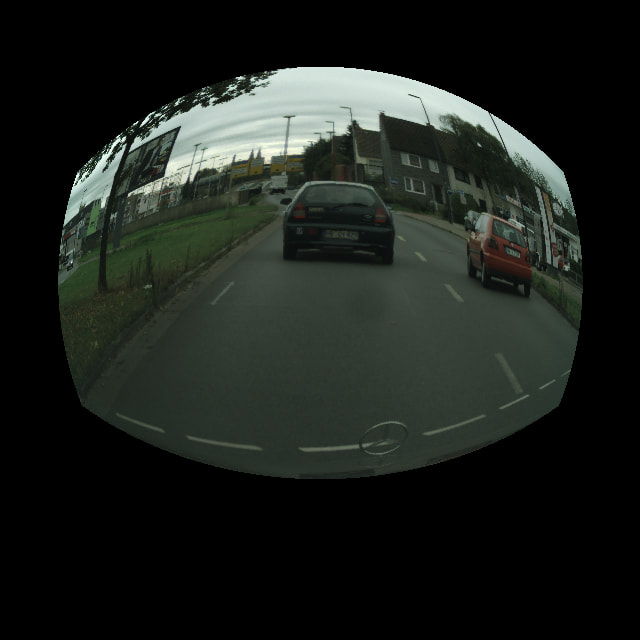}
    \label{fig:roate:down}}
    \subfigure[Camera rotates 15 degree (Z)]
    {\includegraphics[width=0.21\textwidth]{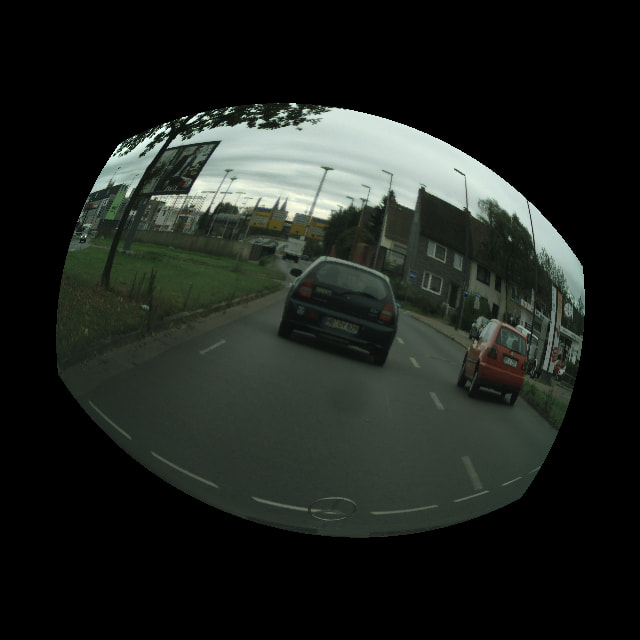}
    \label{fig:rotate:15z}} 
    \subfigure[Camera rotates -15 degree (Z)]
    {\includegraphics[width=0.21\textwidth]{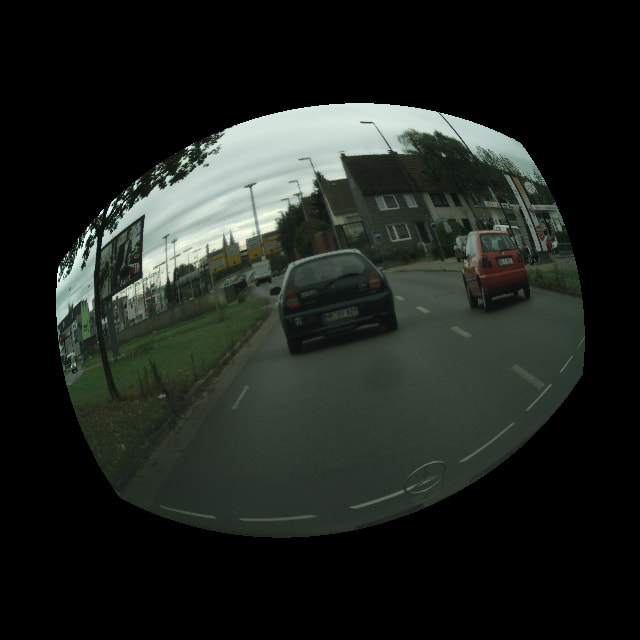}
    \label{fig:rotate:-15z}} \\
    
    \caption{The six DoF augmentation. Except the first row, every image is transformed using a virtual fisheye camera with focal length of 300 pixels. The letter in brackets means that which axis the camera is panning along or rotating around.}
    \label{fig:7D}
    \vspace{0.2in}
\end{figure*}

\section{Method}
\subsection{The principle of seven-DoF augmentation}
The basic principle of data augmentation approach in this paper is to convert the rectilinear image of the world coordinate system into the synthetic fisheye image by using the projection model of the camera and a virtual fisheye camera (see Fig. \ref{fig:PMT}). The so-called seven-DoF augmentation contains the spatial relationship between the world coordinate system and the fisheye coordinate system (six DoF) and the variation in the focal length of the virtual fisheye camera (one DoF). The relative rotation between the two coordinate systems contains three degrees of freedom, and the relative translation between the two coordinate systems also contains three degrees of freedom.

In actual implementation, a point on the fisheye image is mapped to a normal image as follows:

1. For a point $(x_0, y_0)$ in the fisheye plane, it is first mapped to $(x_1, y_1, z_1)$ in world coordinate system using the $r = f\cdot \theta$ principle. ($z_1$ is a relative quantity. We set it to 500 here.)
\begin{equation}
\theta = {\sqrt{x_0^2+y_0^2} \over{f_{fish}}}\label{eq1}
\end{equation}

\begin{equation}
(x_1,y_1) = ({x_0\over{\sqrt{x_0^2+y_0^2}}}tan(\theta), {y_0\over{\sqrt{x_0^2+y_0^2}}}tan(\theta))\label{eq2}
\end{equation}

2. $(x_1,y_1,z_1)$ is mapped to $(x_2, y_2, z_2)$ in pinhole camera coordinate system by a transform matrix.
\begin{equation}
\left(\begin{array}{l}
x_2 \\y_2\\z_2\\1
\end{array}\right)=\left(\begin{array}{ll}
R & t \\0^{T} & 1
\end{array}\right)\left(\begin{array}{l}
x_1 \\y_1\\z_1\\1
\end{array}\right)\label{eq3}
\end{equation}

3. $(x_2, y_2, z_2)$ is mapped to $(u, v)$ in the rectilinear image by a camera Intrinsics. The parameters of cols and rows mean the number of column pixels and row pixels representing the rectilinear image respectively.
\begin{equation}
\left(\begin{array}{l}
u \\v \\1
\end{array}\right)=\frac{1}{z_2}\left(\begin{array}{lll}
z_1 & 0 & cols/2 \\ 0 & z_1 & rows/2 \\ 0 & 0 & 1
\end{array}\right)\left(\begin{array}{l}
x_2 \\ y_2 \\ z_2
\end{array}\right)\label{eq4}
\end{equation}

By following the steps above, we can map every pixel in the fisheye image to the rectilinear image. 
\subsection{Intuitive understanding of seven-DoF augmentation}

\begin{figure*}[tb]
  \centering
    \subfigure[f=200]
    {\includegraphics[width=0.21\textwidth]{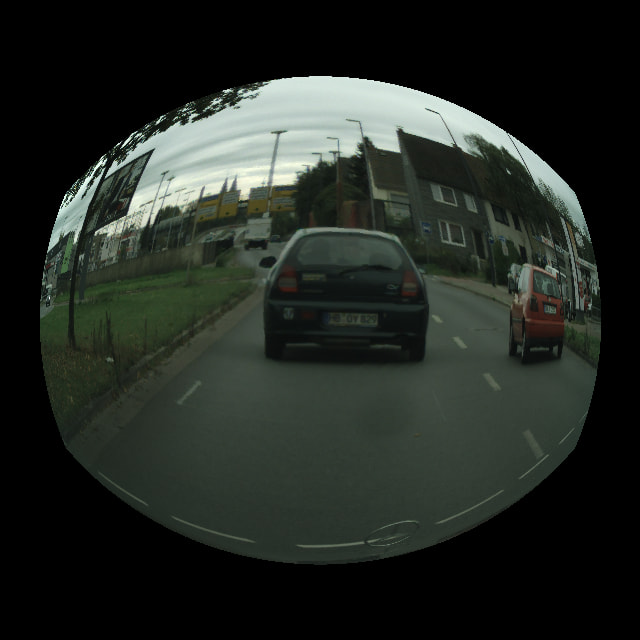}
    \label{fig:zoom:200}}
    \subfigure[f=250]
    {\includegraphics[width=0.21\textwidth]{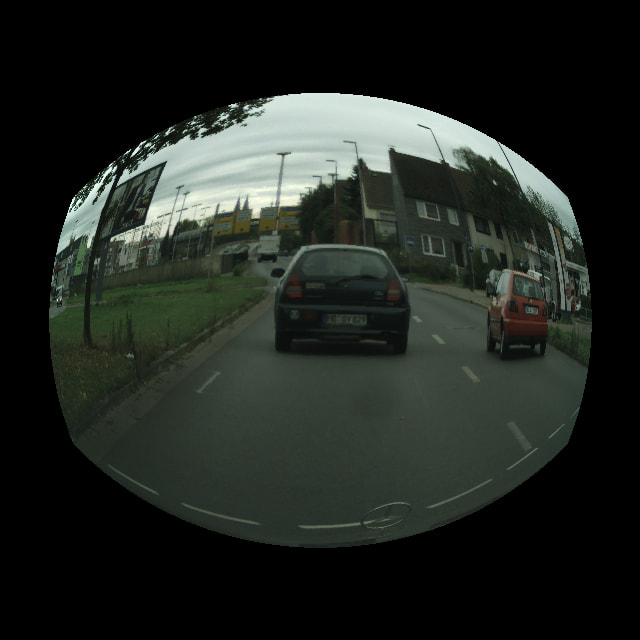}
    \label{fig:zoom:250}}
    \subfigure[f=300]
    {\includegraphics[width=0.21\textwidth]{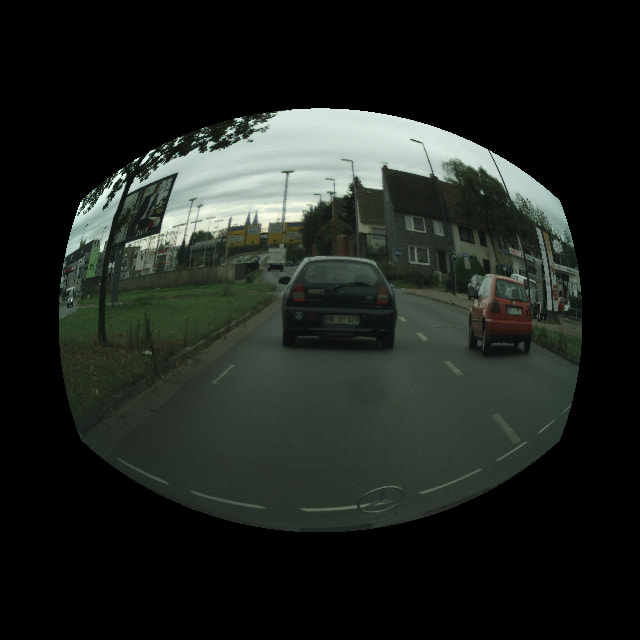}
    \label{fig:trans:RGB}}
     \subfigure[f=350]
    {\includegraphics[width=0.21\textwidth]{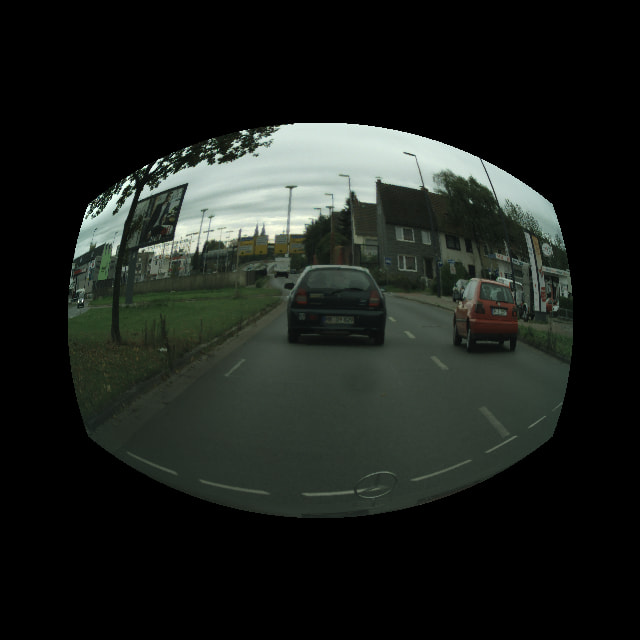}
    \label{fig:zoom:350}} \\
    \caption{the synthetic fisheye images with different f(focal length)}
    \label{fig:focal}
    \vspace{0.2in}
\end{figure*}

This seven-DoF augmentation, while not strictly simulating a fisheye image, can simulate its distortion pattern to some extent. Fig. \ref{fig:7D} illustrates the effects of the six-DoF augmentation. And Fig.\ref{fig:focal} illustrates the virtual fisheye images with different focal lengths.

In the process of training neural networks, we usually use some data augmentation methods to extend the training dataset and reduce overfitting. Data augmentation methods for ordinary rectilinear data include random cropping, random horizontal flipping, panning, rotation, scaling, and color jittering. For fisheye data augmentation based on zoom augmentation, we can use all of these augmentation methods. However, when we use panning or scaling to augment our data, we just augment the data in rectilinear style, not in fisheye style. For a fisheye image, the distortion increases with the distance between the pixel and the center of the image. Besides, for the frames taken by the same fisheye camera, the distortion of the same position of each picture is the same. Therefore, once the fisheye image is translated, the distortion feature of the image will be destroyed.

However, we can find that the seven-DoF augmentation method naturally do the data augmentation in fisheye style.

As shown in Fig. \ref{fig:trans:forward} and Fig. \ref{fig:trans:back}, when we change the relative z-axis position of the virtual fisheye camera coordinate system and the world coordinate system, it simulates the scene of fish-eye camera moving forward and backward. It makes the object closer to the fish-eye camera, which results in the object bigger in the image. The variation of the relative positions of X-axis and Y-axis between the fisheye coordinate system and the world coordinate system actually simulates the position changes of the virtual fisheye camera. Specifically, the augmentation of X-axis translation simulates the changes of the left and right position of the car on the road, while the Y-axis translation simulates the changes of the height of the fisheye camera on the car. It's also understandable that the data augmentation of rotation around three axis can simulate the orientation changes of the fisheye camera.

In practice, the fisheye camera will be placed on a car, and the attitude of the camera is always changing with the time and the turbulence of the car. Also, the position and orientation of the camera will vary from vehicle to vehicle, which results in a different view of the image. However, the neural network is not very good at handling these situation, because it is invariably trained from an existing dataset, and the accuracy of the neural network will decrease if the situation is not similar with the existing dataset. For example, when we use Cityscapes dataset~\cite{cordts2016cityscapes} to train a semantic segmentation network, if we place the camera at a lower position in the actual application, the perspective of the actual image will be different from the training set, which will lead to a decrease in the accuracy. If we have a dataset which contains frames taken by cameras in different orientations and positions, that won't be a problem. But it's a huge project to collect and annotate such a dataset with cameras of different orientations and different positions, especially for fisheye cameras, as fisheye camera has a parameter of focal length and the distortion of fisheye camera varies with focal length. With the seven-DoF augmentation, we can synthesize fisheye images of the camera of different positions, orientations and focal lengths, so that a general semantic segmentation dataset of fisheye camera could be obtained.

\subsection{Comparison of different augmentation methods}
For the seven-DoF augmentation is not generating fisheye images optically, it only simulated fisheye image to some extent. The more dimensions of freedoms it has, the bigger the difference between the synthetic image and the real fisheye scene. We can't directly conclude that seven-DoF is best for fisheye segmentation, so we conducted a series of experiments.

Now we have the following data augmentation for fisheye semantic segmentation: random cropping, random flipping, color jitter, z-aug, six-DoF augmentation and seven-DoF augmentation.For the benchmark, we adopted the data augmentation means of random clipping, random flip, color jitter and fixed f (virtual fisheye camera's focal length), and used the SwiftNet-18 as our semantic segmentation structure. The data augmentation methods are divided into three types: random focal length, random rotation and random translation.

We designed the following data augmentation methods:

1. \textbf{Base Aug}: random clipping + random flip + color jitter + z-aug of fixed focal length

2. \textbf{RandF Aug}: Base Aug + random focal length

3. \textbf{RandR Aug}: Base Aug + random rotation

4. \textbf{RandT Aug}: Base Aug + random translation

5. \textbf{RandFR Aug}: Base Aug + random focal length + random rotation

6. \textbf{RandFT Aug}: Base Aug + random focal length + random translation

7. \textbf{Six-DoF Aug}: Base Aug + random rotation + random translation

8. \textbf{Seven-DoF Aug}: Base Aug + random focal length + random rotation + random translation

\begin{figure}[tb]
  \centering
    
    \subfigure[Results of augmentation methods 1-4.]
    {\includegraphics[width=0.45\textwidth]{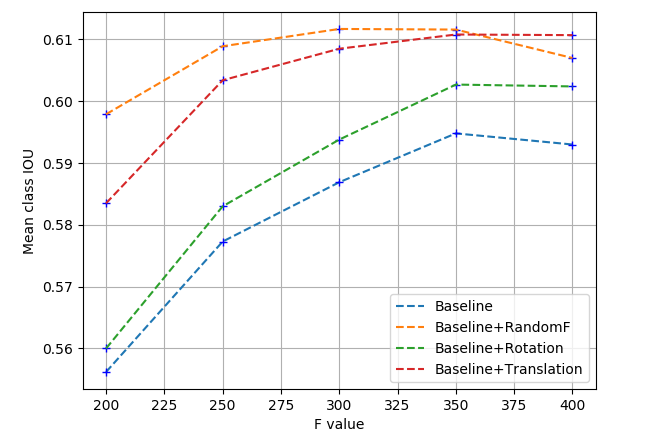}
    \label{fig:dIoU1}} \\
    
    \subfigure[Results of augmentation methods 5-8.]
    {\includegraphics[width=0.45\textwidth]{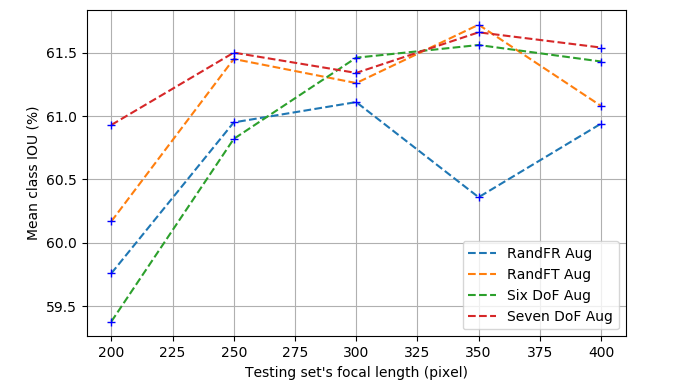}
    \label{fig:dIoU2}}
    
    \caption{Result of all augmentation methods.}
    \label{fig:all-aug}
    \vspace{0.2in}
\end{figure}

First, methods 1 to 4 are tested to compare the performance of a single data augmentation approach (see Fig. \ref{fig:dIoU1}). As it can be seen, when the focal length of the virtual fisheye image of the testing set is larger (the distortion is smaller), the segmentation ability of these models for the distorted image is better. Compared with Base Aug, the RandF Aug, RandR Aug and RandT Aug can evidently improve the accuracy of the model. Moreover, the RandF Aug had the best performance, while RandR Aug made the least improvement on mIoU.

Next, we test the more complex data augmentation methods (see Fig. \ref{fig:dIoU2}). The performance of the model obtained by the combination of multiple data augmentation is better than that of the single method for the testing set of different distortion parameters. It indicates that the seven-DoF Aug achieves the best performance and reaches a high mIoU in every testing sets with different degrees of distortion, which proves the robustness of the seven-DoF augmentation. However, the six-DoF Aug has a worst performance compared with other approaches. The previous experiment already shows that random focal length improves the mIoU most, while random rotation improves the mIoU least. Therefore, it’s understandable that the six-DoF Aug without random focal length augmentation performs the worst.

To sum up, the performance of different data augmentation methods are shown in Table \ref{table:tab1}. While the fixed z-aug performs worst, the seven-DoF augmentation we proposed almost performs best in all testing datasets. Compared with other augmentation methods, the seven-DoF has significant advantages, especially in testing datasets with larger distortion (smaller $f$ value).

\begin{table*}[tb]
\caption{performance of different data augmentation}
\label{table:tab1}
\begin{center}
\begin{tabular}{c c c c c c}
\hline
Data augmentation & mIoU ($f=200$) & mIoU ($f=250$) & mIoU ($f=300$) & mIoU ($f=350$) & mIoU ($f=400$)\\
\hline
Fixed z-aug & 0.5562 & 0.5773 & 0.5869 & 0.5948 & 0.5930 \\
Random z-aug & 0.5979 & 0.6089 & 0.6117 & 0.6116 & 0.6070\\
Six-DoF aug & 0.5938 & 0.6082 & \textbf{0.6146} & 0.6156 & 0.6143\\
Seven-DoF aug & \textbf{0.6093} & \textbf{0.6150} & 0.6134 & \textbf{0.6166} & \textbf{0.6154}\\
\hline
\end{tabular}
\end{center}
\end{table*}

\subsection{Hyper-parameters settings}
The above experiments demonstrate the effectiveness of the seven-DoF data augmentation method. However, arbitrarily setting the parameter values of different degrees of freedom cannot maximize the superiority of the method. In some cases, the value of a parameter is too unreasonable and may even cause the accuracy of the model to decrease. In order to set our hyper-parameters more scientifically, we conduct several experiments to test the model with different values of the hyper-parameters.

\begin{figure}[tb]
  \centering
    
    \subfigure[Results of different translation range along x-axis]
    {\includegraphics[width=0.45\textwidth]{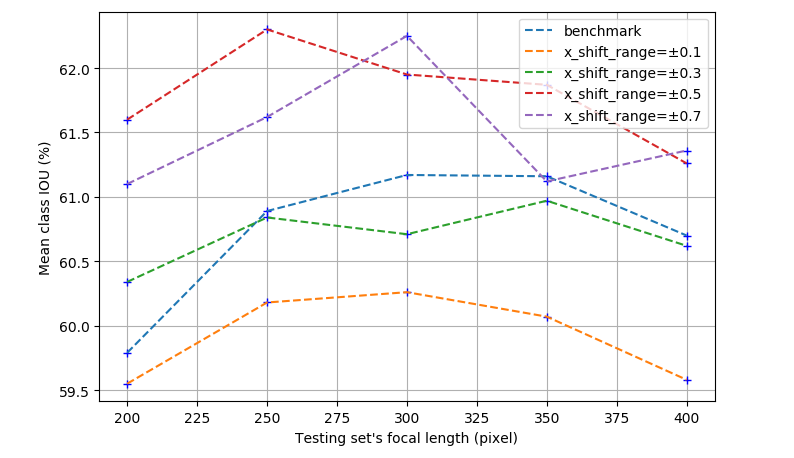}
    \label{fig:x_shift_exp}} \\
    
    \subfigure[Results of different rotation range around y-axis]
    {\includegraphics[width=0.45\textwidth]{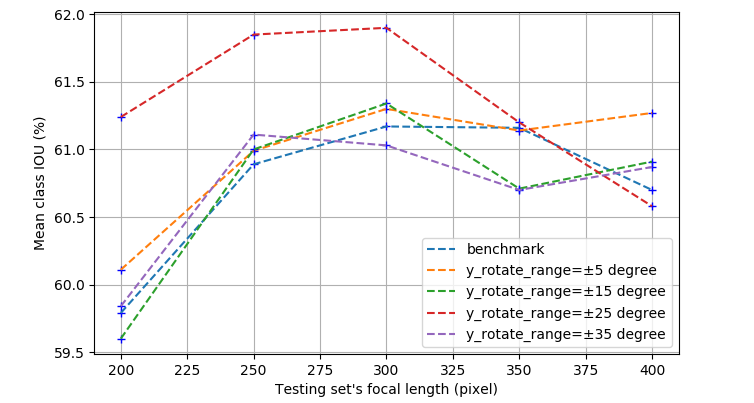}
    \label{fig:y_rotate_exp}}
    
    \caption{Result of hyper-parameters settings.}
    \label{fig:param}
    \vspace{0.2in}
\end{figure}

The first experiment explores the setting of the translation parameters. The translation parameters include translations along the x, y, and z axes. The translation along the x and y axes have similar effects on image distortion. We first discuss translation along the x-axis. We do not directly set an absolute pixel value as the variation range of the translation parameter, but uses a normalized value v of [0,1] to represent the translation. In the code implementation, we set $v * fish\_width$ to the pixel value of the camera coordinate system's final translation, where $fish\_width$ is the width of the virtual fisheye image finally generated. Taking the experiment with $f$ value variation range of [200, 400] as the benchmark, the range of the camera coordinate system translation along the x axis is set to [-0.1, 0.1], [-0.3, 0.3], [-0.5, 0.5], [-0.7, 0.7] respectively. The results reveal that the model works best when $v = [-0.5,0.5]$ (see Fig. \ref{fig:x_shift_exp}).

For the translation range v along the y-axis, by analogy, it should also be set to [-0.5, 0.5]. However, considering that our application scenario is urban autonomous driving, the actual meaning of the camera coordinate system translation along the y-axis is the height variation of the fisheye camera. In practice, when the vehicle is driving in different lanes, the left and right positions of the camera may vary greatly, but the height of the camera does not change much, even for different models of cars. Therefore, the parameter v for translation along the y-axis is set to [-0.1, 0.1]. For the parameter v of translation along the z axis, in the code implementation, we normalize it to (-1, 1), and the actual translation distance is the parameter v multiplied by the focal length of the pinhole camera. Here we set it to [-0.4, 0.4]. Under this parameter range, the distortion of the virtual fisheye image will not be too much.

Similarly, for the setting of the rotation parameters, based on the experiment where the $f$ value variation range is [200, 400] as the benchmark, the variation range of the fisheye camera coordinate system rotation around the y axis is set to [-5, 5], [-15, 15], [-25, 25], [-35, 35] degrees respectively. The results (Fig. \ref{fig:y_rotate_exp}) indicate that the model performs best when the rotation parameter range is set to [-25, 25]. For the parameter range of the camera coordinate system rotating around the x axis, we also set it to [-25, 25] degrees. For the rotation parameter setting around the z-axis, the effect it produces is the rotation of the fisheye image that is ultimately generated. We set it to [-25, 25] degrees.

\begin{figure*}[tb]
  \centering
  
    \subfigure[]
    {\includegraphics[width=0.2\textwidth]{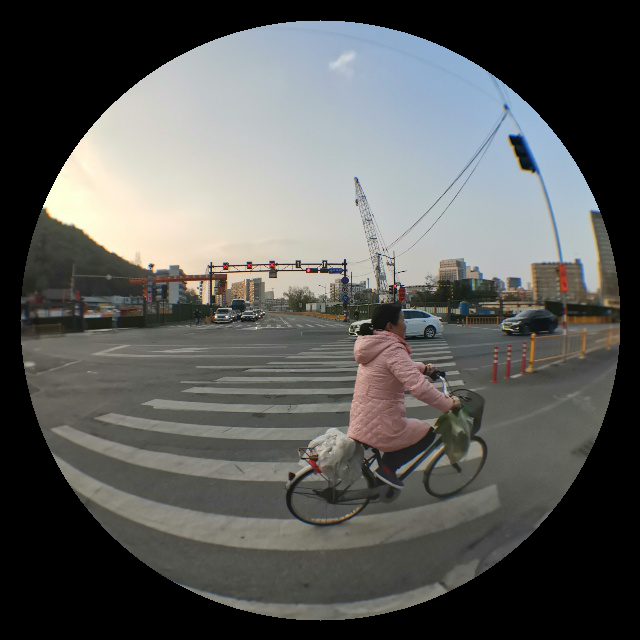}
    \includegraphics[width=0.2\textwidth]{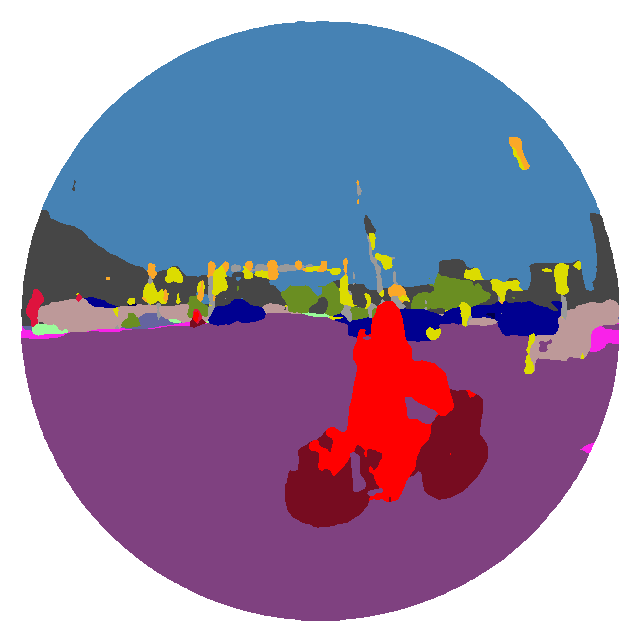}
    \label{fig:real:1}}
    \subfigure[]
    {\includegraphics[width=0.2\textwidth]{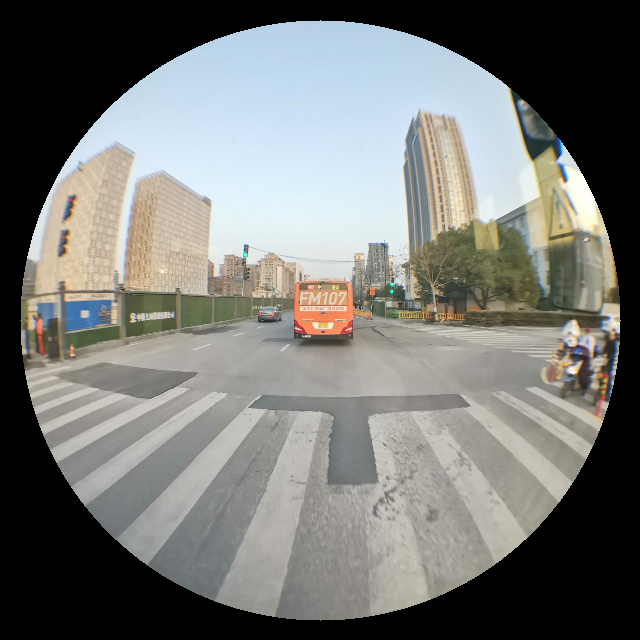}
    \includegraphics[width=0.2\textwidth]{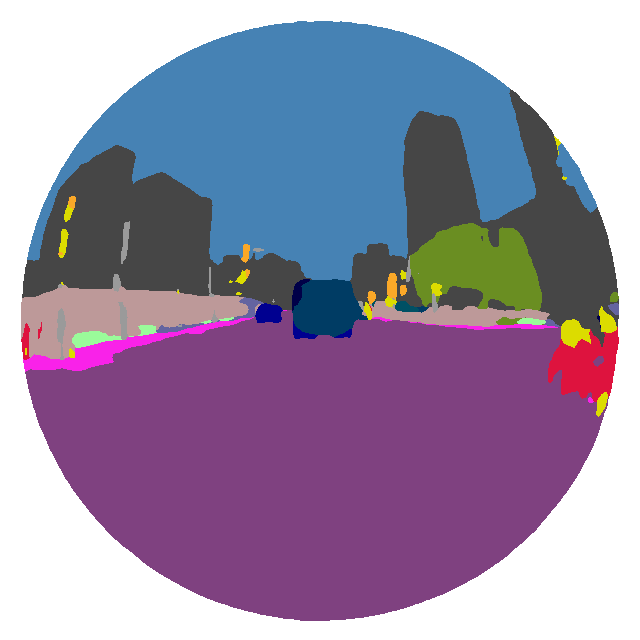}
    \label{fig:real:2}} \\
    
    \subfigure[]
    {\includegraphics[width=0.2\textwidth]{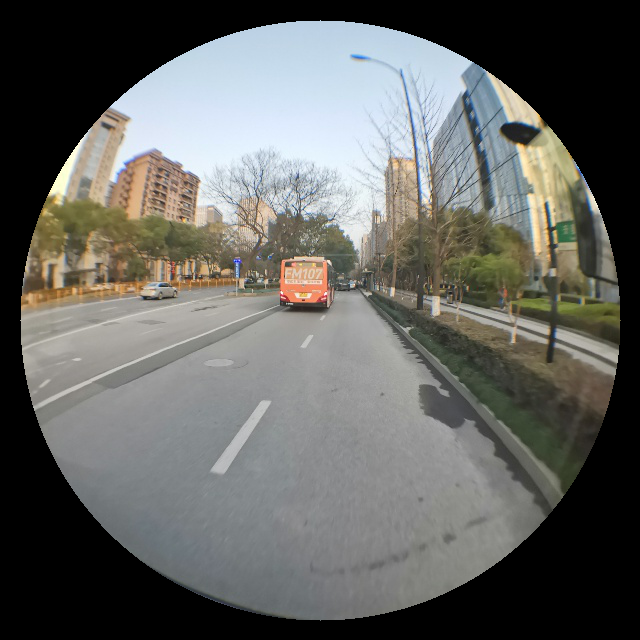}
    \includegraphics[width=0.2\textwidth]{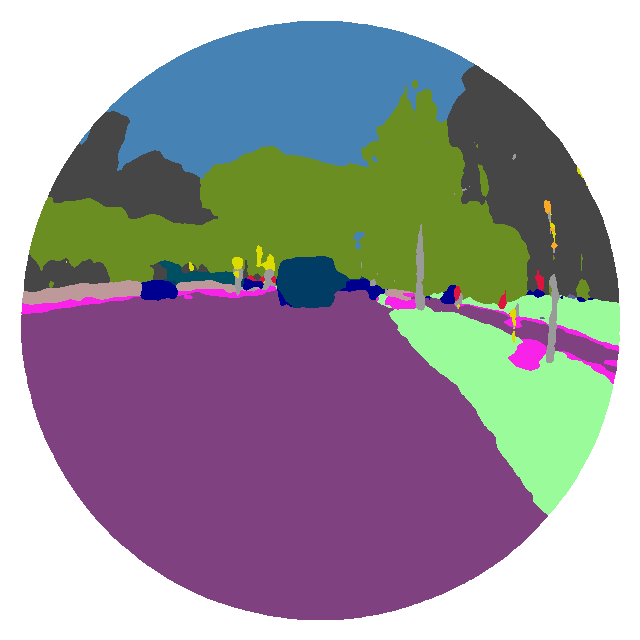}
    \label{fig:real:3}}
    \subfigure[]
    {\includegraphics[width=0.2\textwidth]{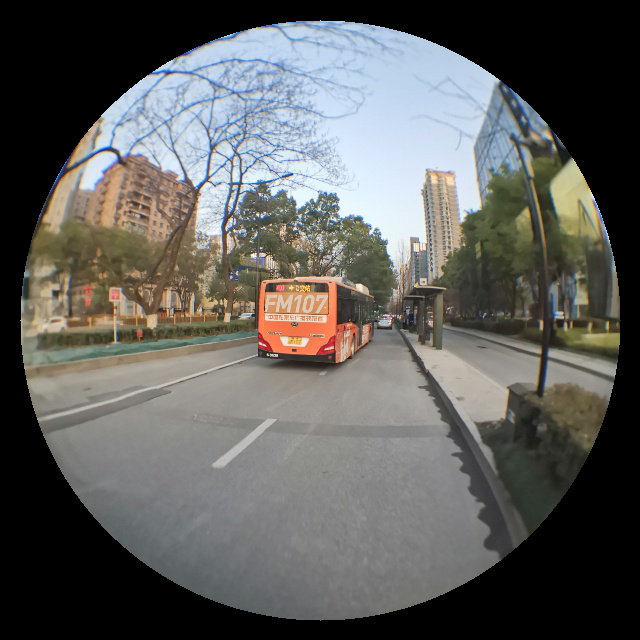}
    \includegraphics[width=0.2\textwidth]{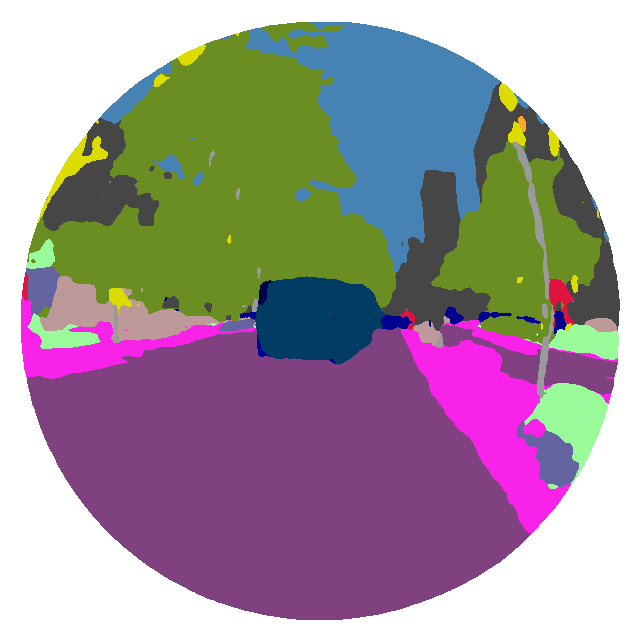}
    \label{fig:real:4}} \\
    
    \subfigure[]
    {\includegraphics[width=0.2\textwidth]{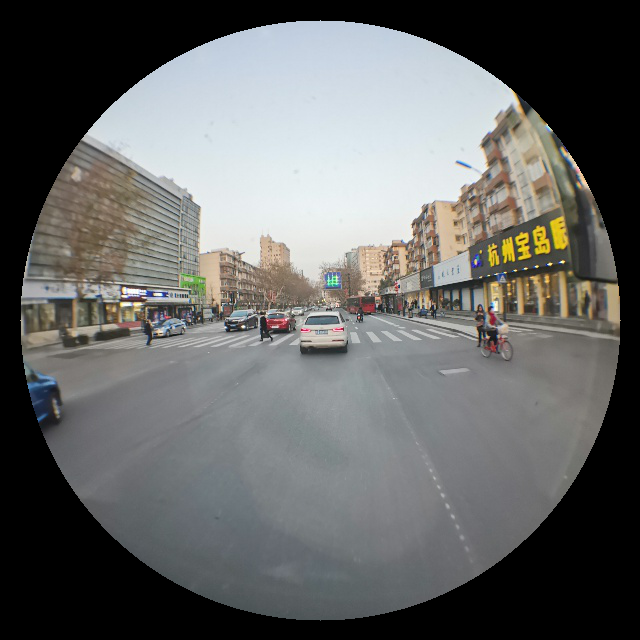}
    \includegraphics[width=0.2\textwidth]{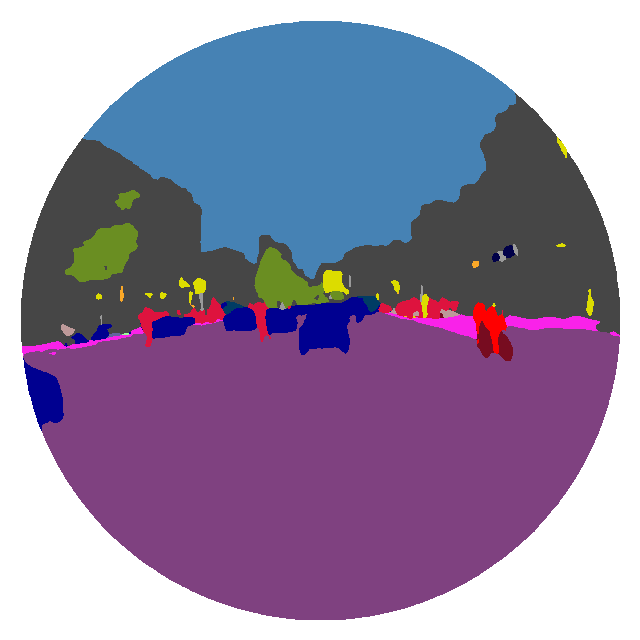}
    \label{fig:real:5}}
    \subfigure[]
    {\includegraphics[width=0.2\textwidth]{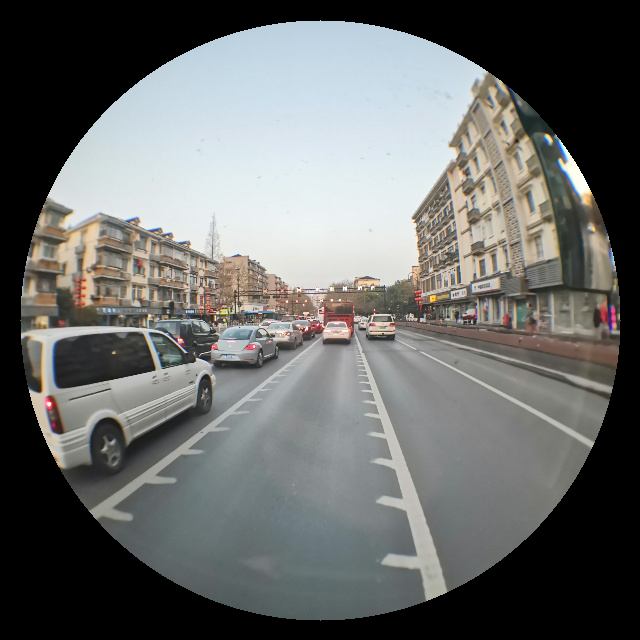}
    \includegraphics[width=0.2\textwidth]{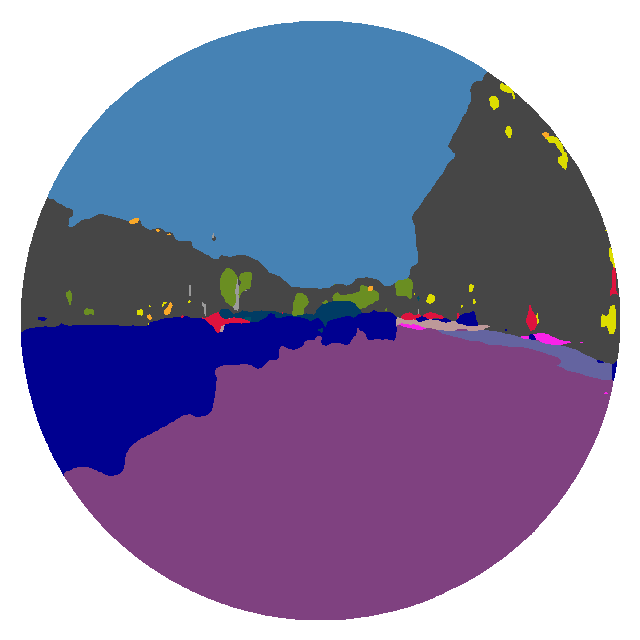}
    \label{fig:real:6}} \\
    
    \subfigure[color legend]
    {\includegraphics[width=0.8\textwidth]{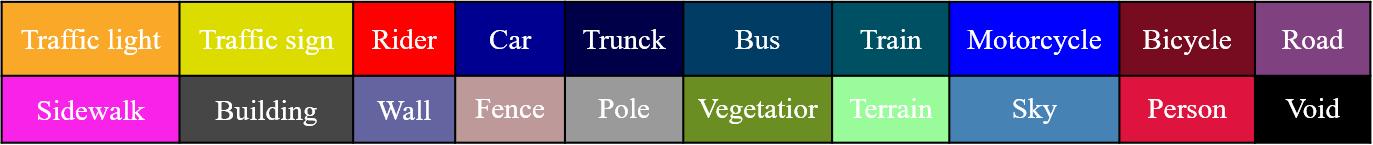}
    \label{fig:real:color}}
    
    \caption{Semantic segmentation of real fisheye images.}
    \label{fig:real}
    \vspace{0.2in}
\end{figure*}

\section{Experiments}
\subsection{dataset and CNN structure}
CityScapes dataset~\cite{cordts2016cityscapes} is a well-known dataset in the field of autonomous driving. It was recorded in street scenes from 50 different cities, and provides 5000 finely annotated frames, in addition to a larger set of 20000 coarsely annotated frames. Within the 5,000 pixel-level annotated frames, 2,975 frames were used for training, 500 frames for validation, and 1,525 frames for testing. We used the 2,975 training data and 500 validation data to conduct our experiments.
\subsubsection{Training dataset}
We directly use the 2975 dataset as our raw dataset, with the method of online data augmentation to transform the rectilinear data to fisheye images to train our neural networks. The original training set is the rectilinear image of 1024*2048 pixels. After data augmentation, we unified them into fisheye images of 640*640 pixels. This paper uses the method of online augmentation, that is, the parameters of seven-DoF augmentation change in every batch. The advantage of this method is that each image of the training set is transformed into a different fisheye image each time it is fed to the semantic segmentation network (the data augmentation part contains random parameters), which can greatly increase the richness of the training set.

\subsubsection{Testing dataset}
Testing set is to use the z-aug (zoom augmentation) to transform the 500 pieces of rectilinear cityscapes' validation data into virtual fisheye data. Different focal lengths are used to generate testing sets for a better evaluation of our models. We generate five testing sets and their focal length is 200, 250, 300, 350, 400 respectively. If the model has superior generalization performance, it should perform well on all testing sets.

\subsubsection{CNN structure and training details}
As this work focuses on the data augmentation, we simply choose SwiftNet-18~\cite{orsic2019defense} (a lightweight CNN structure with ResNet 18 as it’s backbone), which has a U-net~\cite{Ronneberger2015U} structure, to conduct our experiment. We use the SwiftNet-18 with ImageNet pre-training. Just the same as the paper\cite{orsic2019defense}, the pre-trained parameters are updated with Adam optimizer with learning rate of $1\cdot10^{-4}$. The learning rate decays with cosine annealing to the minimum value of $2.5\cdot10^{-5}$. And the other parameters are updated with 4 times bigger learning rate and 4 times bigger weight decay. We utilize the focal loss\cite{lin2017focal} as the loss function of the semantic segmentation. Batch size is set to 12 and we train for 200 epochs on Cityscapes. We choose the last epoch's parameters as the final model.

\subsection{Real fisheye image test}
To test the generalization performance of our model, we collected fisheye images of real urban street scenes. For convenience, we adopted an external fisheye lens for a mobile phone with a field of view of about 180 degrees and clip it to a smartphone. On the bus, we held the smartphone equipped with the external fisheye lens. As the bus navigated, we collected a series of fisheye image data of urban street scenes. We resized the obtained images to 640 * 640 resolution, and applied our model (based on seven-DoF augmentation) to the obtained images. Fig. \ref{fig:real} depicts the segmentation performance of our model in different scenes. As it can be seen, basically all categories are well segmented.

\section{Conclusions And Future Work}
This paper proposes a general virtual fisheye data augmentation method, the seven-DoF augmentation. This method transforms a rectilinear dataset into a fisheye dataset in a comprehensible way, synthesizing fisheye images taken by cameras with different orientations, different positions, and different $f$ values, which significantly improves the generalization performance of fisheye semantic segmentation. It provides a universal semantic segmentation solution for fisheye cameras in different autonomous driving applications. In addition, even if you already have a fisheye dataset, this method is still very meaningful. Because in practice, it is unlikely that the training set will have the same parameters as the real installed fisheye lens. The distortion parameters of different fisheye lenses are different, and the fisheye images obtained by different parameters such as orientation and camera height are also different. The dataset taken by a fisheye camera with fixed parameters cannot be well adapted to segmentation task for images taken by cameras with other different parameters.

This paper also discusses the setting of hyper-parameters for data augmentation. The parameters specially designed for urban autonomous driving scenarios evidently improves the segmentation accuracy of the model. Besides, this article proposes a convenient method to obtain fisheye images, which combines a smartphone and an external fisheye lens. Finally, when applied to real fisheye images, our model achieves precise segmentation results. 

This article mainly focuses on the data augmentation method of fisheye semantic segmentation, and does not design the network structure according to the characteristics of fisheye images. In the future, we plan to make some CNN structural improvements for fisheye images specially. In addition, data augmentation methods for real fisheye datasets are also a promising research direction.


\bibliographystyle{IEEEtran}
\bibliography{bib.bib}

\end{document}